\documentclass[11pt,a4paper]{article}

\usepackage{algorithm}
\usepackage{algorithmic}
\usepackage{pslatex}
\usepackage{tightenitemize}
\usepackage[pdftex]{graphicx}
\newcommand\hide[1]{}
\newcommand{\figref}[1]{Fig.~\ref{fig:#1}}
\newcommand{\secref}[1]{Sec.~\ref{sec:#1}}
\usepackage[hyperref]{acl2019}
\usepackage{times}
\usepackage{latexsym}
\usepackage{mathtools}
\usepackage{relsize}
\usepackage{textcomp}

\usepackage{url}

\aclfinalcopy 
\def\aclpaperid{***} 

\title{Relating Word Embedding Gender Biases to Gender Gaps: \\ 
A Cross-Cultural Analysis}
\author{Scott Friedman, Sonja Schmer-Galunder, Anthony Chen, and Jeffrey Rye\\
  SIFT, Minneapolis, MN USA \\\
  \texttt{\{friedman, sgalunder, achen, rye\}@sift.net}}
\begin{document}
\maketitle
\begin{abstract}
Modern models for common NLP tasks often 
employ machine learning techniques and train on journalistic, social
media, or other culturally-derived text.
These have recently been scrutinized for racial and gender biases,
rooting from inherent bias in their training text.
These biases are often sub-optimal and recent work poses methods
to rectify them; however, these biases may shed light on
actual racial or gender gaps in the culture(s) 
that produced the training text, thereby helping us understand cultural context through big data.
This paper presents an approach for quantifying
gender bias in word embeddings, and then using them to characterize 
statistical gender gaps in education, politics, economics, and health. 
We validate these metrics on 2018 Twitter data spanning 51 U.S. regions 
and 99 countries.
We correlate state and country word embedding biases with 18 international and 5 U.S.-based statistical gender gaps, characterizing regularities and predictive strength.
\end{abstract}

\section{Introduction}
\label{sec:introduction}

Machine-learned models are the \emph{de facto}
method for NLP tasks.
Recently, machine-learned models that utilize \emph{word embeddings} (i.e., vector-based representations of word semantics)
have come under scrutiny for biases and stereotypes, e.g., in race and gender, arising primarily from biases in their training data \cite{bolukbasi2016man}. 
These biases produce systematic mistakes, so recent work has developed
\emph{debiasing} language models to improve NLP models' accuracy
and and remove stereotypes \cite{zhao2018gender,zhang2018mitigating}.

Concurrently, other research has begun to characterize how biases
in language models correspond to disparities in the cultures that produced the training text, e.g., by mapping embeddings to survey data \cite{kozlowski2018geometry},
casting analogies in the vector space to compute that ``man is to woman as doctor is to nurse'' \cite{bolukbasi2016man}, or varying the training text over decades and mapping each decade's model bias against its statistical disparities to capture periods of societal shifts \cite{garg2018word}.

Building on previous work,
this paper presents initial work characterizing word embedding biases with
 statistical \emph{gender gaps} (i.e., discrepancies in opportunities and status 
across genders).
This is an important step in approximating cultural attitudes and relating them to cultural behaviors. 
We analyze 51 U.S. states and 99 countries, by (1) training separate word embeddings for each of these cultures from Twitter and (2) correlating the biases in these word embeddings with 5 U.S.-based and 18 international gender gap statistics.

Our claims are as follows:
(1) some cultural gender biases in language are associated with gender gaps;
(2) we can characterize biases based on strength and direction of correlation with gender gaps; and
(3) themed word sets, representative of values and social constructs, capture different dimensions of gender bias and gender gaps.

We continue with a brief overview of gender gaps (\secref{gender-gaps}) and then a description of our training data (\secref{training-data}) and four experiments (\secref{experiments}).
We close with a discussion of the above claims and future work (\secref{conclusions}).

\section{Gender Gaps and Statistics}
\label{sec:gender-gaps}
Within the social sciences, anthropologists often attempt to explain the asymmetrical valuations of the sexes across a range of cultures
with respect to patterns of social and cultural experience \cite{rosaldo1974woman}.  
This work contributes to this research by updating traditional qualitative approaches with computational methods.

The public sphere is often associated with male and agents traits (assertiveness, competitiveness) in domains like politics and executive rules at work. 
Private or domestic domains linked to family and social relationships are traditionally related to women, although social relationships are considered more important by people independent of gender \cite{friedman2000work}. 
Gender gaps arise form these asymmetrical valuations, e.g., where men are typically over-represented and have higher salaries compared to women \cite{mitra2003establishment, vincent2013women, bishu2017systematic}.

We utilize diverse gender gap statistics in this work.
For international data, we use 18 gender gap metrics comprising the Global Gender Gap Index (GGGI) originally compiled for
the World Economic Forum's 2018 Gender Gap Report.\footnote{http://reports.weforum.org}
The GGGI measures clearly-defined dimensions for which reliable data in most countries was available \cite{hawken2013cross}.
For domestic data, we use a 2018 report from the U.S. Center for Disease Control (CDC) on male and female exercise rate \cite{blackwell2018state}, wage gap and workforce data published by the U.S. Census Bureau in 2016, female percentages of math and computer science degrees from Society of Women Engineers,\footnote{http://societyofwomenengineers.swe.org/} and female percentages of each state's legislators from Represent Women's 2018 Gender Parity Report.\footnote{http://www.representwomen.org}

\section{Training Data}
\label{sec:training-data}

Our training data include public tweets from U.S. and international 
Twitter users over 100 days throughout 2018, including the first ten days of each of the first ten months.
We use tweet's location property to categorize by location, and we include only English tweets in our dataset.

We filtered out all tweets with fewer than three words, and following other Twitter-based embedding strategies \cite[e.g.][]{li2017data}, we replaced URLs, user names, hashtags, images, and emojis with other tokens.
We divided the processed tweets into two separate datasets: (1) U.S. states and (2) countries.
This helps us validate our approach with multiple granularities and datasets.

The international dataset contains 99 countries with varying number of tweets, ranging from 98K tweets (Mauritius) to 122M tweets (U.K).
The U.S. states dataset contains 51 regions (50 states and Washington, D.C.) ranging from  450K tweets (Wyoming) to 65M tweets (California).
For both datasets, we sampled 10 million tweets for all cultures that exceeded that number.
These corpora are orders of magnitude smaller than other approaches for tweet embeddings \cite[e.g.,][]{li2017data}.

We use Word2Vec to construct word vectors for our experiments, but we compare
Word2Vec with other algorithms in our analyses (\secref{algo-exp}).

\begin{figure*}
  \centering
  \includegraphics[width=\textwidth]{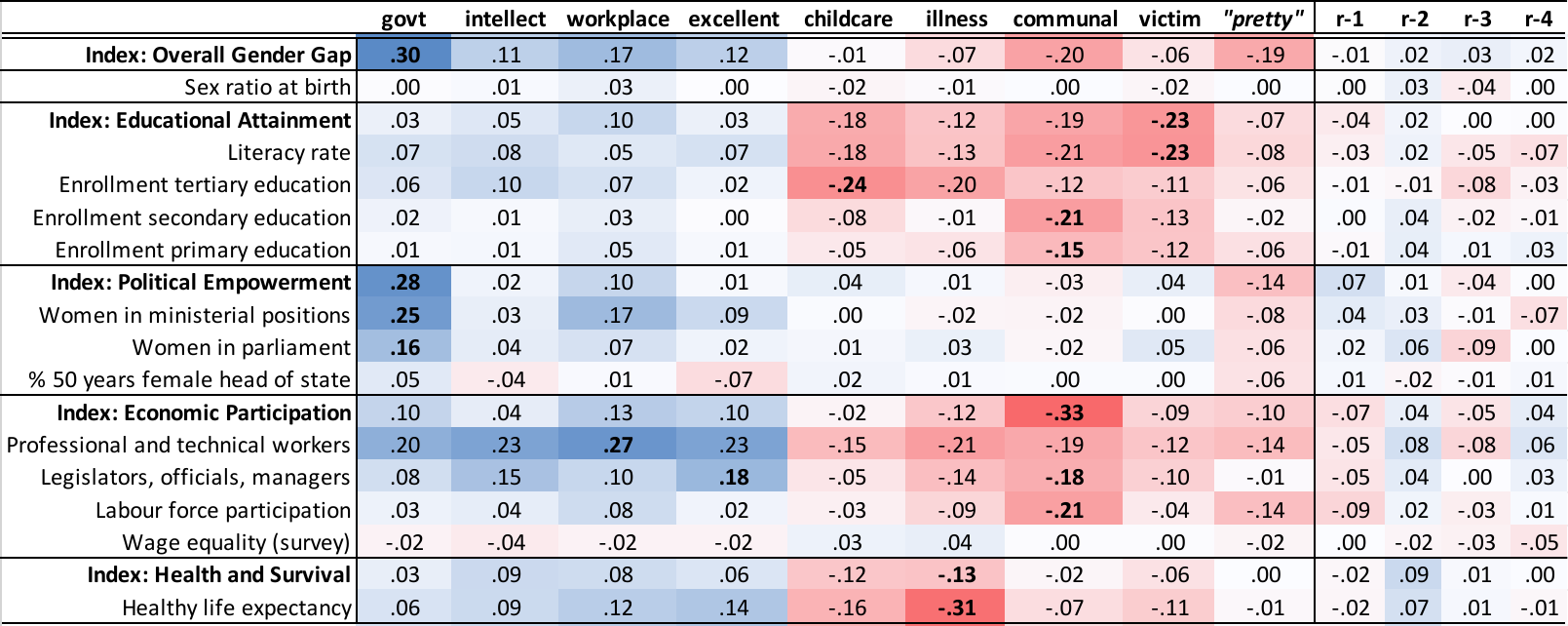}
  \caption{Correlation of themed neutral word sets' gender bias (columns) against categories of gender gaps from worldbank.org (rows).
  Values are $R^2$ coefficient of determination, where negation is added to indicate inverse correlation.
  The rightmost four word sets (\emph{r-1 to r-4}) were randomly sampled from the vocabulary
  for comparison.
  }
  \label{fig:intl-correlations}
\end{figure*}

\section{Experiments}
\label{sec:experiments}
\hide{
We describe our empirical results, including an international data analysis (\secref{intl-exp}), a U.S. data analysis (\secref{domestic-exp}), a comparative analysis of word embedding algorithms (\secref{algo-exp}), and analyses of the valence and dominance of words that correlate with U.S. gender gaps (\secref{sensitivity-exp}).
}
\begin{figure}[h]
  \centering
  \includegraphics[width=\linewidth]{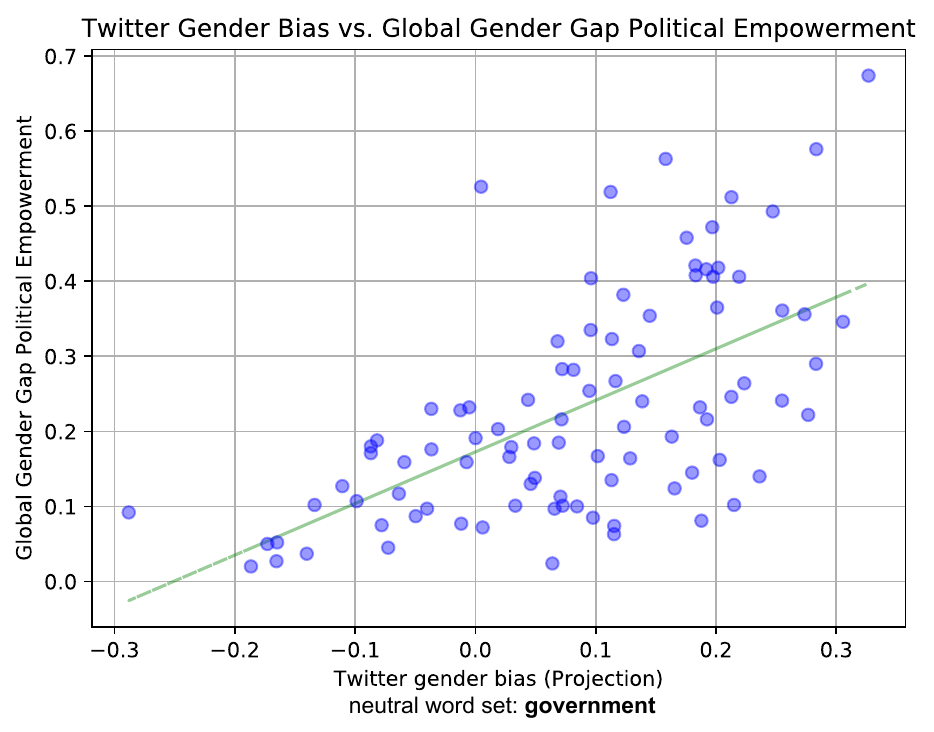}
  \caption{Correlation of country's gender bias of government words (x-axis; female association increases in positive direction) against the World Economic Forum's political empowerment gender gap index (y-axis; gender gap decreases in positive direction). 
  }
  \label{fig:intl-plot}
  \vspace{-.1in}
\end{figure}

\subsection{International Analysis}
\label{sec:intl-exp}

Our international and U.S.-based analyses have an identical experimental setup, varying only in the gender gap statistics and the word embeddings.

Our materials included word-sets based in part on survey data \cite{williams1990sex} and recent work on word embeddings \cite{garg2018word}.
These word-sets included (1) \emph{female words} including female pronouns and nouns, (2) \emph{male words}, including male pronouns and nouns, and (3) \emph{neutral words} that were grouped thematically.
For instance, we used \emph{appearance} and \emph{intellect} adjectives from \cite{garg2018word}, and we generated other thematic word sets representative of social constructs: \emph{government} (democrat, republican, senate, government, politics, minister, presidency, vote, parliament, ...), \emph{threat} (dangerous, scary, toxic, suspicious, threat, frightening	 ...), \emph{communal} (community, society, humanity, welfare, ...), \emph{criminal} (criminal, jail, prison, crime, corrupt, ...), \emph{childcare} (child, children, parent, baby, nanny, ...), \emph{excellent} (excellent, fantastic, phenomenal, outstanding, ...) and others.

We use the same male and female word sets for international and U.S. state analyses, and we compute per-gender vectors $\overrightarrow{female}$ and $\overrightarrow{male}$ by averaging the vectors of each constituent word, following \cite{garg2018word}.
For any country or state's word embedding, we compute the \emph{average axis projection} of a neutral word set $W$ onto the male-female axis as:
\begin{equation}
avg_{w \in W}\mathlarger{(}{\overrightarrow{w} \cdot
    \frac{\overrightarrow{female} - \overrightarrow{male}}{||\overrightarrow{female} - \overrightarrow{male}||_{2}}}\mathlarger{)}
\end{equation}
This average axis projection is our primary measure of gender bias in word embeddings.

For any neutral word list (e.g., government terms), we compute the average axis projection for all countries (or states) and compute its correlation to international (or U.S.) gender gaps.
\figref{intl-plot} plots each country's government/political word bias against the World Economic Forum's Political Empowerment Gender Gap sub-index (from 0 to 1, where greater score indicates less gap).
The value 0.0 on the x-axis indicates no gender bias, and female bias increases along the x-axis.

Consequently, \figref{intl-plot} is consistent with the hypothesis that--- globally, over our set of 99 countries--- women's political influence and power increase (relative to men) as political language shows a more female bias.

We present results of each thematic word set regressed against all available international statistics.
For each pair of themed word set and gender gap statistic, the algorithm (1) performs feature selection on 20\% of the countries to optionally down-select from the set of words in the themed word set, (2) uses the down-selected word set to compute the $R^2$ determination against the full set of countries, and then (3) repeats a total of five times and averages the answers.
Feature selection monotonically increases the $R^2$, and using 20\% of countries helps prevent over-fitting.

\begin{figure*}
  \centering
  \includegraphics[width=\textwidth]{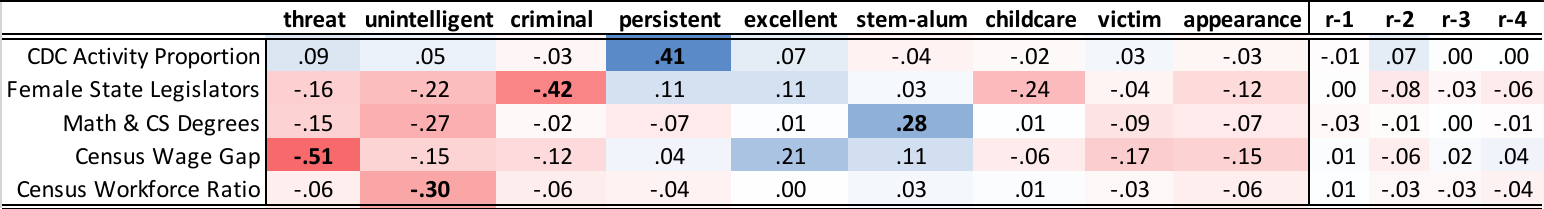}
  \caption{Correlation of themed neutral word sets' gender bias (columns) against U.S. gender disparity statistics from CDC, US Census Bureau, and Represent Women's 2018 Gender Parity Report (rows).
  Values are $R^2$ coefficient of determination, where negation is added to indicate inverse correlation.
  The rightmost four word sets (\emph{rand 1-4}) were randomly sampled from the vocabulary
  for comparison.
  }
  \label{fig:state-correlations}
  \vspace{-.1in}
\end{figure*}

\figref{intl-correlations} includes our results over this analysis, grouping gender gap sub-indices (bold) with their related statistics.
This illustrates that different word sets vary in their correlation direction and strength across different statistic groups: the \emph{political} set is positively correlated with the political empowerment subgroup and marginal on some economic statistics, but weak over health and education; intellectual and workplace terms positively correlate with economic statistics but are weak predictors otherwise; \emph{illness} terms indirectly correlated with health and survival statistics, but are weak correlates elsewhere; and so-forth.
The word \emph{``pretty,''} shown in \figref{intl-correlations}, was the single word with the strongest determination against the overall gender gap and other sub-indices.
\figref{intl-correlations} also includes four randomly-generated word sets, which do not exceed $R^2 = 0.09$ for any gender gap.

The selective correlation of these thematic word sets with related gender gap statistics supports our claim that gender biases in word embeddings can help characterize and predict statistical gender gaps across cultures.
Since we trained our embeddings on tweets alone--- with as few as 98K tweets for some countries--- this also supports our claim that social media is a plausible source to compute a culture's gender bias in language.

None of our themed word sets strongly correlated with: (1) sex ratio at birth, which was 1.0 for the vast majority of countries; (2) percentage of last 50 years with female head of state; and (3) survey-based wage equality.
The latter two gender gaps may correlate with other themed word sets, or they
may have a more complex or nonlinear relationship to a culture's gender bias in language.

\begin{figure}
  \centering
  \includegraphics[width=\linewidth]{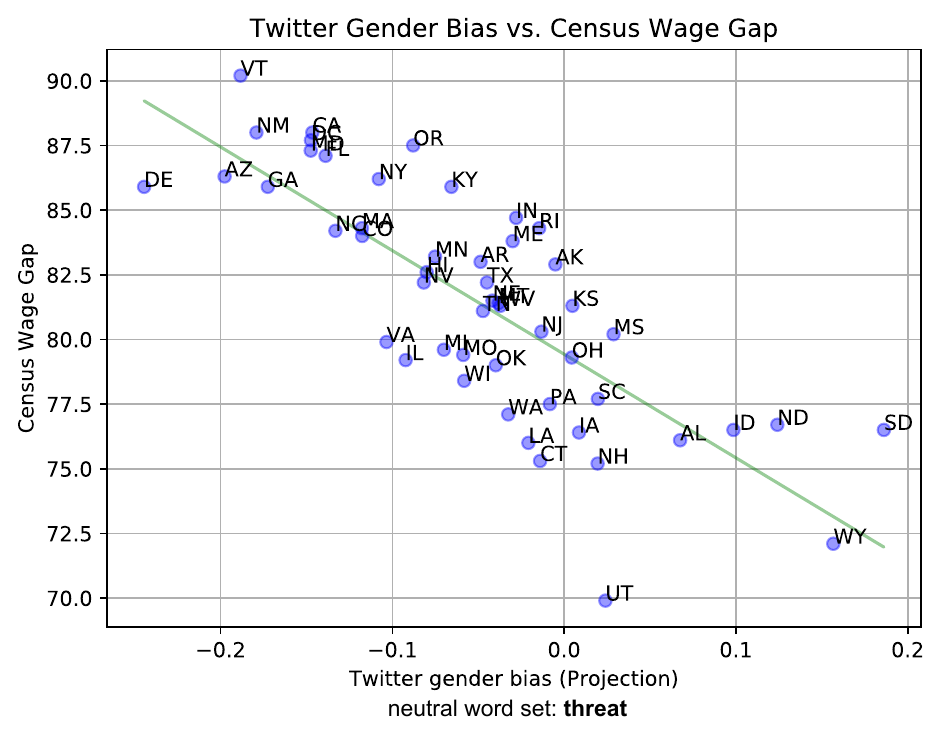}
  \caption{Correlation of states' gender bias of threat words' (x-axis; female association is positive direction) against the pay gap reported by U.S. Census Bureau in 2016 (y-axis; pay gap decreases in positive direction). 
  }
  \label{fig:us-plot}
  \vspace{-.1in}
\end{figure}

\subsection{U.S. State Analysis}
\label{sec:domestic-exp}

Our analysis of 51 U.S. regions (50 U.S. states and Washington, D.C.) is analogous to our \secref{intl-exp} international analysis; we only vary the word embeddings and the statistical gender gap data.

\begin{figure*}[ht]
  \centering
  \includegraphics[width=.8\textwidth]{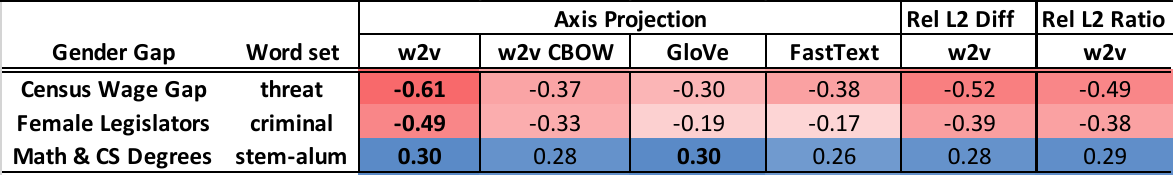}
  \caption{Comparison of three bias metrics and four word embedding algorithms correlating themed word sets' gender bias with U.S. gender gap statistics.
  Unlike in \figref{intl-plot}, we perform feature selection using all countries.}
  \label{fig:method-compare}
  \vspace{-.1in}
\end{figure*}

\figref{us-plot} shows an example of indirect correlation ($R^2 = 0.51$) of our \emph{threat} word set (threat, dangerous, toxic, suspicious, scary, frightening, horrifying, ...) against U.S. Census Bureau data reported in 2016 on the gender pay gap.
The y-axis indicates cents on the dollar earned by women for the same work as men, ranging from 69.9\textcent~(UT) to 91.2\textcent~(VT).
This inverse correlation is consistent with the hypothesis that when masculinity is threatened in some cultures, men react by asserting dominance \cite{zuo2000breadwinner, schmitt2001good}.

\figref{state-correlations} illustrates different word sets' determination on U.S. regions' statistical gender gaps.
The word set describing persistence and devotion had strongest direct correlation with reduced gender gap in exercise.
The word set for criminal behavior had strongest negative correlation with female proportion of state legislators.
Words for STEM disciplines and alumni directly correlated with increased percentages of female math and CS degrees.
Threat-based words negatively correlated with pay equality, and words for unintelligent and inept negatively correlated with female percentage of the workforce.
Other word sets from the international analysis (e.g., childcare and victimhood) had less determination of gender gaps than in the international setting.

As with our international analysis, this domestic analysis supports our claim that gender biases in cultural language models can predict and characterize statistical gender gaps.

\begin{figure}
  \centering
  \includegraphics[width=.9\linewidth]{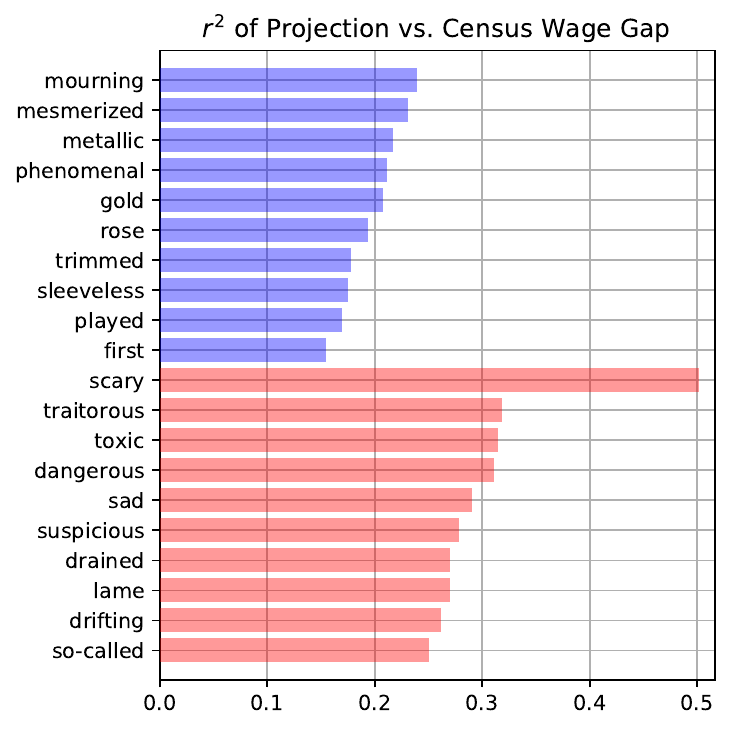}
  \caption{Ten adjectives with highest bias correlations to reduced pay gap (top, blue), and ten with highest correlation to increased pay gap (bottom, red).
  }
  \label{fig:pay-gap-adj}
\end{figure}

\subsection{Algorithm Comparison}
\label{sec:algo-exp}

We compare four word embedding algorithms and three bias metrics using our gender gap statistics and word sets.
We compare four algorithms: (1) GloVe, (2) Word2Vec (skip-gram), (3) CBOW Word2Vec, and (4) FastText (skip-gram). 
\hide{
Each of these algorithms constructs a vector of each word based on the occurrence that word has with other words. This is held in a large sparse matrix in which each row represents a word and each column is a context and so each element is the frequency a word occurs in a given context. 
}
For each algorithm we utilize a window  size 10, filter words that occur fewer than 5 times, and produce 200-dimension output vectors. 
\hide{and dimensionality of each vector set to d=200. Distance between two word vectors x and y or their semantic similarity is determined by the cosine similarity $cos(\pmb x, \pmb y) = \frac {\pmb x \cdot \pmb y}{||\pmb x|| \cdot ||\pmb y||}$.
}

\textbf{GloVe} \cite{pennington2014glove} uses count-based vectorization to reduce dimensionality by minimizing reconstruction loss.
The dot product of two GloVe vectors equals the log of the number of times those two words occur near each other.

\textbf{Word2Vec} \cite{mikolov2013word2vec} uses a predictive model to learn geometric encodings of words through a feed-forward neural network optimized by stochastic gradient descent. 
The Word2Vec \emph{continuous bag-of-words} (CBOW) setting predicts the most probable word given a context.
The Word2Vec \emph{skip-gram} setting differs slightly by inputting a target word and predicting the context.

\textbf{FastText} \cite{joulin2016fasttext} characterizes each word as an n-gram of characters rather than an atomic entity. So each word vector is the sum of word vectors of the target word's n-gram (e.g. ``app,'' ``ppl,'' ``ple'' for ``apple''). This is especially useful for rare words that might not exist in the corpus and accounting for misspellings.

\figref{method-compare} illustrates the above word embedding algorithms used
on three different correlated word sets and statistics.
In addition to comparing different word embedding algorithms, we also compare three different bias metrics on the Word2Vec algorithm: (1) the \emph{axis projection} metric defined in \secref{intl-exp}; (2) the \emph{relative L2 norm difference} \cite{garg2018word}; and the (3) \emph{relative L2 norm ratio}.
Unlike the axis projection, metrics (2) and (3) both compute the L2 norm from each word in the neutral word set to the $\overrightarrow{male}$ and $\overrightarrow{female}$ vectors, and then subtract or divide the two norms, respectively, returning the average over the word set.

The \figref{method-compare} results demonstrate that the gender bias is present in the product of all four word embedding algorithms, and is detectable with all three metrics.
The Word2Vec approach with axis projection yields the highest coefficient of determination--- for both direct and indirect correlation--- across all three gender gap and word set pairs.
This is the algorithm and bias metric that we use for all other experiments.

\begin{figure}
  \centering
  \includegraphics[width=\linewidth]{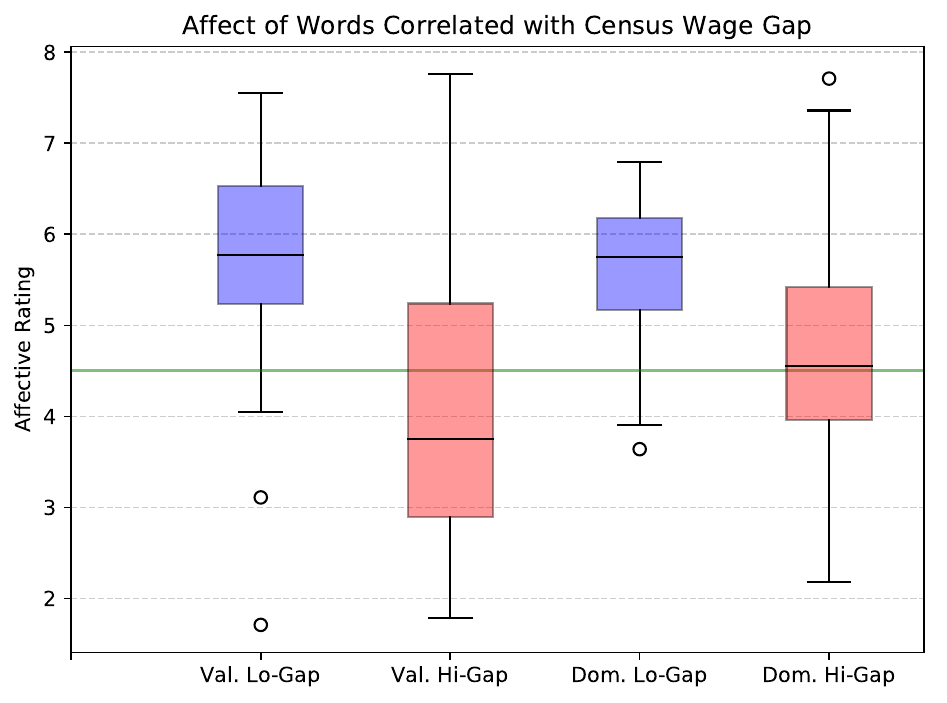}
  \caption{Valence and dominance scores for decreased pay gap words (blue) and increased pay gap words (red).
  Affect is neutral at 4.5 (plotted in green).
  }
  \label{fig:pay-gap-affect}
\end{figure}

\subsection{Valence and Dominance Analysis}
\label{sec:sensitivity-exp}

Our \secref{intl-exp} and \secref{domestic-exp} experiments specified word sets 
\emph{a priori}, but we can also identify and analyze the individual words whose
gender biases directly and indirectly correlate with statistical gender gaps
to find trends and commonalities.

We identified all adjectives in the word embeddings using WordNet and then computed each adjective's $R^2$ score for direct or indirect correlation with each U.S. gender gap statistic.
We filtered down the adjectives to those that correlate directly or indirectly with $R^2 > 0.1$.
To illustrate, \figref{pay-gap-adj} plots ten highest $R^2$ words for direct (blue) and indirect (red) correlation against the pay gap, where blue adjectives' female bias correlates with \emph{reduced} pay gap (higher wages) and red adjectives' female bias correlates with \emph{increased} pay gap (lower wages) in U.S. embeddings.

For each statistic, we measured the \emph{valence} and the \emph{dominance} of the directly- and indirectly-correlated adjectives using scores from \citet{warriner2013norms}.
\figref{pay-gap-affect} shows a box plot of the valence and dominance of the reduced-gender-gap adjectives (blue) against increased-gender-gap adjectives (red) for the gender pay gap statistic, where the valence and dominance values for reduced gap (\emph{Lo-Gap}) are significantly higher than the valence and dominance for increased gap (\emph{Hi-Gap}) via t-test, where $p < 1.0e^{-7}$.

The same valence and dominance pattern held for adjectives directly and indirectly correlated with economic and educational gaps (i.e., Census Workforce Ratio, Female State Legislators, and Math \& CS Degrees), where the valence and dominance of Lo-Gap words were significantly higher than Hi-Gap words with $p < .005$ throughout.
The difference in valence and dominance for CDC Activity gap was not significant.

\section{Conclusions}
\label{sec:conclusions}

This paper characterized gender biases in Twitter-derived word embeddings from multiple cultures (99 countries and 51 U.S. regions) against statistical gender gaps in those cultures (18 international and 5 U.S.-based statistics).

We demonstrated that thematically-grouped word sets' gender biases correlate with gender gaps intuitively: word sets with a central topic or valence correlate with gender gaps of a similar topic, in a meaningful (positive or negative) direction.
This supports our claims (from \secref{introduction}) that (1) cultural biases in language are correlated with cultural gender gaps and (2) we can characterize biases based on strength and direction of correlation with these gaps.
We also demonstrated that these correlations are selective: not all topical word sets' biases correlate with all gender gaps, and random word sets do not correlate.  
This supports our claim that themed word sets capture different dimensions of gender bias and gender gaps.

Finally, we identified adjectives whose biases were highly correlated with increased and decreased gender gaps in education and economics, and we found that the adjectives correlated with \textit{increased} gender gaps had statistically significantly lower valence and dominance than those correlated with \textit{decreased} gender gaps.
This is evidence of a cross-cutting attitude towards gender that we can characterize with future work.

The results of our three bias analyses are consistent with the social theory that differences in implicit gender valuation (e.g., linguistic gender bias) manifest in different gender opportunities and status (e.g., gender gaps) \cite{berger1972status,rashotte2005gender}.
Specifically, when a culture attributes greater competence and social status to a gender, that gender receives higher rewards and evaluations \cite{dini2017second}.

\paragraph{Limitations and Future Work.}

Our use of English-only tweets facilitated comparison across embeddings, but it eliminates the native language of many countries and creates cultural blind-spots.
Specifically, our use of English tweets does not capture the voices of those that (1) lack access to technology, (2) have poor knowledge of English, and (3) simply do not use Twitter.
One might even argue that the gender bias effects may be even more pronounced off-line due to social desirability effects.
Expanding to other languages presents additional challenges, e.g., gendered words and many-to-one vector mappings across languages, but recent language transformers facilitate this \cite{devlin2018bert}. 
Incorporating additional languages and cultural texts are important next steps.

Previous Twitter word embedding approaches blend tweets with news or Wikipedia to improve NLP accuracy, using orders of magnitude more text per embedding \cite{li2017data}.
Blending tweets with news may improve the embeddings' accuracy for NLP tasks, but it also risks diluting their implicit biases.

Finally, 
while our analyses illustrate correlations between gender biases and statistical gender gaps, they do not describe causality and they have limited interpretive power.
We believe that integrating these methods with additional data and causal models (e.g., Dirichlet mixture models and Bayesian networks) will jointly improve interpretation and accuracy.

\section*{Acknowledgments}
This research was supported by funding from the Defense Advanced Research Projects
Agency (DARPA HR00111890015). The views, opinions and/or findings expressed 
are those of the authors and should not be interpreted as representing the official views 
or policies of the Department of Defense or the U.S. Government.

\bibliography{acl2019}
\bibliographystyle{acl_natbib}

\end{document}